# Beyond Interpretable Benchmarks: Contextual Learning through Cognitive and Multimodal Perception

Nick DiSanto
California Baptist University, nick.c.disanto@gmail.com

*Abstract* – With state-of-the-art models achieving high performance on standard benchmarks, contemporary research paradigms continue to emphasize general intelligence as an enduring objective. However, this pursuit overlooks the fundamental disparities between the high-level data perception abilities of artificial and natural intelligence systems. This study questions the Turing Test as a criterion of generally intelligent thought and contends that it is misinterpreted as an attempt to anthropomorphize computer systems. Instead, it emphasizes tacit learning as a cornerstone of general-purpose intelligence, despite its lack of overt interpretability. This abstract form of intelligence necessitates contextual cognitive attributes that are crucial for human-level perception: generalizable experience, moral responsibility, and implicit prioritization. The absence of these features yields undeniable perceptual disparities and constrains the cognitive capacity of artificial systems to effectively contextualize their environments. Additionally, this study establishes that, despite extensive exploration of potential architecture for future systems, little consideration has been given to how such models will continuously absorb and adapt to contextual data. While conventional models may continue to improve in benchmark performance, disregarding these contextual considerations will lead to stagnation in human-like comprehension. Until general intelligence can be abstracted from task-specific domains and systems can learn implicitly from their environments, research standards should instead prioritize the disciplines in which AI thrives.

## 1 INTRODUCTION

While machine learning systems have proven beneficial in various industrial applications, they lack a comprehensive ability to generalize in abstract environments. While heuristically performative, these models are examples of Artificial Narrow Intelligence (ANI): models strictly bound to task-oriented domains [1, 2]. This is a notable contrast to human perception, the driving factor behind abstract thought and metacognition [3, 4]. Artificial General Intelligence (AGI), which represents complete knowledge abstraction, is the theoretical embodiment of these attributes. However, attributes of ANI are pervasive in all contemporary AI products, with multidisciplinary products struggling to generalize across applications [5].

In order to deconstruct this problem, it is necessary to isolate the applications of ANI and AGI. This requires a distinction between interpretable intelligence and general-purpose intelligence. While intelligent thought is a notably conjectural concept, the AI community conventionally considers it to be the ability to apply learned structures to specific tasks [6, 7]. An important consideration of this definition is the requirement of interpretable benchmarks and task-specific data preparation. General intelligence, on the other hand, is not necessarily illustrative at benchmark performance. It can instead be considered a domain-independent ability to learn from any data and generalize it in a variety of abstract environments [8, 9].

As recent work has attempted to establish criteria for AI to approach human-level intelligence (HLI), it does not acknowledge abilities that are unachievable in the current research paradigm. While AGI is currently theoretical, current shortcomings are unavoidable when compared to natural human comprehension. Thus, this study seeks to establish the history and goal of general intelligence, the concrete distinctions between natural and artificial learning architectures, and the steps necessary to build a model capable of tacit learning.

## 2 THE TURING TEST AND GOAL OF AGI

HLI is a natural benchmark for AGI, but the Turing Test (TT) was the most notable work to formally establish it. Conceived by Alan Turing in 1950 [10], the "Imitation Game" involves a human communicating with a hidden entity. To pass the test, a machine must render the human incapable of distinguishing between other human users and AI imposters. Turing's goal in establishing this experiment was simply to create a tangible benchmark for AGI, considering "thought" has been arbitrarily defined since before Descartes [11].

As one of the pioneering endeavors to characterize an artificial system with human-like qualities, the TT laid the foundation for artificial neural networks and other conventional architectures. Notably, these demonstrations established the objective of AI development as "matching human capability," which would persist rigidly for decades. This dangerous rhetoric burdens the TT with several limitations. For starters, it suggests that humans are the only reasonable standard for measuring an intelligent system—a misconception that will be explored in greater depth later. Its efficacy also hinges on the aptitude of its participants [12] and presupposes that language serves as the best measure of intelligent thought [13]. This assertion is contested by Dreyfus [14], who establishes that human learning is mainly tacit and unquantifiable. The TT also implies that high performance in a performative task, such as conversation, is a surefire demonstration of rational thought. However, as established in the definition of AGI, general intelligence should not be contingent on interpretable benchmark performance. This is illustrated by the "Chinese Room" argument [15]: a correct response is not always a thoughtful one.

The strict binarity of the TT also demonstrates the issue of simply trusting model output. The TT culminates with a simple "yes" or "no" answer regarding the computer's aptitude, with the system remaining a complete black box. This fails to reveal its cognitive process, its approach to approximating human-like reasoning, or, in the case of failure, its proximity to success. The distinction between an individual with mathematic equations memorized and one who can manually derive them illustrates this point. Merely acknowledging a correct answer provides no transparency into problem-solving capacity.

With a rejection of the Turing Test's general-purpose utility, it becomes imperative to establish a new goal for AGI. Instead of conforming to prevailing research trends, this objective should align with the capabilities of modern learning architectures. Unfortunately, this has not stayed consistent; whether spurred by trendy developments in the field, portrayals of the media [16], or Turing himself, research has persisted in the pursuit of AGI. However, while industry applications of ANI have expectations, AGI is struggling to match the general aptitude of even a young child. The inability for cutting-edge products to exhibit striking cognitive abilities prompt questions about the practicality of such approaches.

Russell & Norvig [17] analogize AI development to the aeronautical engineering research that led to successful artificial flight. The most viable aeronautic solutions did not come by modeling the source – pigeons. Instead, the general principles that govern birds were used as inspiration to build a machine with a different, but equally viable, structure. Similarly, success in AI ventures does not need to come by perfectly imitating human thought but by simply reaching effective solutions.

## 3  BIOLOGICAL AND PERCEPTUAL DISPARITIES

The distinctions between natural and artificial data perception serve as a foundation for the possibilities of advanced intelligent systems. Distinguishing both the individual and shared attributes can establish the feasibility of mitigating these differences and bridging the gap.

As previously mentioned, artificial learning systems exhibit striking similarities, as they are designed to replicate the learning capacity of the human brain by simplifying data and training through abstraction. Grossberg [18] provides a particularly notable example, building models inspired by specific regions of the brain. Likewise, de Garis [19] employs simplified neural networks based on cellular automata to simulate the transmission of growth instructions within a three-dimensional space.

The fidelity of these architectural models demonstrates that the emulation of human brain functionality is not only feasible but underway. Consequently, the distinction between humans and AI cannot be attributed to internal data processing. This solicits an important question: why is the gap between AI and HLI so substantial when their underlying processes are so similar? The answer lies in sensory perception and information-gathering abilities. The notion of AGI is built upon the hasty assumption that the deep and comprehensive data essential for HLI is readily accessible to machines. However, this assumption is fundamentally flawed; the external domains in which these systems operate are entirely incomparable. AGI's obstacle is not its underlying structure but its ability to process contextualized information.

While their effectiveness may vary, all biological systems share advanced perceptive data acquisition and learning mechanisms. Illeris [20] elucidates human learning as a dual-phase process consisting of an interaction between the learner and his or her environment, followed by an elaborative psychological process. Simply put, the ability to unconsciously absorb information and utilize it for future decision-making is fundamentally biological. Humans often remain unaware of these gradual ideological shifts due to their extensive data collection, derived from their continuous daily experiences. This inherent mechanism is why humans instinctively gravitate towards pleasure and avoid pain with little deliberation [21]. In the aggregate, these interactions are a constant engagement with internal and external surroundings [22], building the implicit intuition that guides human decision-making. Consequently, human experience can be likened to a continuous, dynamic dataset with an infinite range of potential values to explore.

In contrast, accumulating data for a supervised model demands deliberate efforts, meticulous data organization, and intentional preprocessing. The model is also only exposed to values that have been deemed relevant by an annotator. This makes its training data entirely contingent on human decision-making and restricts its ability to arrive at subconscious insights. Given that computers lack the capacity to "experience the world," attempting to apply human-modeled learning architectures to artificial datasets is comparing apples and oranges with respect to data perception. These differences are examined by Hoyes [23], who contends that the critical component that computers are missing is the inability to instantiate three-dimensional perceptions from two-dimensional sense modalities. The human brain should primarily be regarded as a highly efficient data consuming machine.

While ANI can undoubtedly surpass human performance in specific specializations, generalizing a machine's knowledge across an infinitely diverse domain is impractical. Instead of striving to emulate human intelligence, modeling interactivity offers a more pragmatic approach, emphasizing abstract learning capabilities. A noteworthy example illustrating these challenges can be found in Vinyals & Le's neural conversational model [24]. How do even high-budget language understanding models struggle to grasp the nuances of human language, despite their brute-force learning methods? The performance disparity can be attributed not to architectural differences but rather to the context of the data they encounter.

## 4   THE LIMITATIONS OF CONTEXTUAL COGNITION

Each of the billions of daily human experiences triggers complex nerve signals in the body, transmitting highly intricate information to the brain. The brain can then rapidly ascertain the source, significance, and contextual relevance of these sensory inputs. This vast amount of harmoniously processed data yields an extraordinary depth of understanding. Conversely, optimizing AI proves challenging due to its inability to grasp contextual nuances, stemming from its narrow scope and its incapacity to perceive relative implications.

Humans excel as sophisticated learners because their information encompasses three essential contextual components: **generalizable experience, emotion and moral responsibility, and implicit prioritization.**

### 4.1 – Generalizable Experience

The most pivotal contextual tool humans possess is the ability to generalize past experiences to new situations. This capability stands as a linchpin of intelligent life, shaping the rationality of existence. Without robust generalization abilities, it could be argued that machines remain entirely unable to reason. Chollet [25] offers a perspective on intelligence as a system's "skill-acquisition efficiency," which can be seen as an alternative lens through which to perceive generalization, given that deducing unwritten instances can be an efficient means of acquiring input [26].

While AI can exhibit a degree of generalization, it often struggles with common-sense reasoning and questions of inference [27]. Lacking real-world context renders straightforward conclusions much more complex.

The subsequent descriptions outline two distinctive methods by which humans generalize previous experiences, along with their contrasts in AI.

#### 4.1.1 – Predicting Future Activity From Past Events

Humans have an uncanny ability to predict not only the hedonic consequences of predictable activities but also events that have yet to occur [28]. By utilizing previously established rules, such as the laws of physics, humans can infer what they anticipate will transpire and render the outcome seemingly straightforward. This capacity is often referred to as "metacognition," a skill that equips humans to navigate consistent uncertainties. In contrast, AI starts from a clean slate with each unique situation. This severely limits its capabilities, as a substantial portion of its computational power is expended in confirming what humans can intuitively piece together.

#### 4.1.2 – Multimodal Association

Humans are remarkably adept at drawing deep connections among seemingly unrelated elements. Consider a juror in a courtroom, for instance: they can assess a suspect's testimony, eyewitness accounts, and physical evidence, considering each factor to arrive at a reasoned verdict. Their adaptability stems from processes that can interact synergistically. Conversely, while multimodal learning has gained recent traction [29], a machine would be quite an ineffective juror, as can be seen in empirical suspect analysis applications. While networks specifically designed to classify suspects based on a forensic sketch may be moderately accurate [30], they are unable to comprehend other aspects of the suspect. This limitation arises from their incapacity to apply their knowledge in fundamentally unique environments, even if it merely requires a single logical inference. Even if several distinct models excel in each of these individual tasks, interrelating them to reach a broader and contextually meaningful solution remains an insurmountable challenge.

### 4.2 – Emotion and Moral Responsibility

The inability to perceive emotions remains a significant facet of AI's limited cognitive capabilities—a factor that also excludes it from being considered a "conscious" entity [31]. While emotional responses are not inherently necessary for enhancing a system's accuracy or predictability, they play a role in every human decision. Many complex problems, such as those in politics or religion, are unavoidably rooted in individual subjectivity. Even if this can introduce bias and be potentially harmful, emotions constitute a fundamental aspect of the human experience. Since machines cannot interpret personal experiences and emotions, they lack a personal philosophy to guide them through intricate real-world situations. Therefore, for AGI to aspire to human-like tendencies, it must also demonstrate the capacity to expressively engage with the world around it.

Similarly, every human decision is rooted in the individual's inherent values and moral responsibilities. In order to arrive at an informed decision, humans draw from a lifetime of experiences to discern their moral compass, allowing for abstract problem-solving in unseen situations. To illustrate, a fiscal conservative unfamiliar with the details of a new market regulation can likely infer, based on their principles, that they are going to disagree with it. Conversely, machines are constrained to their training data, making subjective discernment much more challenging. This also contributes to the hesitance of many researchers to trust models to make significant decisions. The "gut feeling" that humans often rely on when making choices enables them to navigate ethical dilemmas with relative ease, while AI tends to falter when confronted with these intangible issues [32].

*4.3 – Implicit Prioritization*

Humans possess a remarkable ability to imbue information with prioritized meaning, a crucial instrument by which they navigate situations [33]. On the contrary, AI relies on brute force data collection to contextually understand situations. While the single dimensionality of these applications can often sidestep complications, human experiences are imbued with external implications that alter their impact.

To illustrate, consider two men: Tom watching a comedy movie and Jim attending his father's funeral. Both activities may take an hour, yet Jim undeniably extracts more profound meaning from his experience than Tom. The human brain possesses the extraordinary ability to discern what events should hold substantial consequences for future decisions. As Voss [34] explains it, reality presents far more information than can be reasonably processed, making an automated filtering system imperative.

AI, on the other hand, has no internal metric with which to prioritize events. While all activities may play a role in the learning process, they will be weighed wildly improperly. Until they can comprehend the profound implications implied by different situations, artificial systems will struggle to distinguish the events that deserve focused attention.

## 5 DISCUSSION

Conventional research posits that general intelligence is the ability to improve without an extensive knowledge base. However, this study argues differently: a significant amount of knowledge is indeed essential for general intelligence. The crucial distinction instead lies in the lack of anthropomorphized data collection processes. This disparity leads to an unavoidable conclusion regarding the future of AI research: proponents of AGI have been pursuing the wrong objective for decades. The fundamental contextual differences between these systems necessitate distinct applications. Humans and AI should concentrate on the domains for which they are most adept: abstract problem-solving and domain-specific tasks, respectively.

This consideration will also be pivotal in shaping future sentiments toward AI research since the rapid expansion of the field has begun to instill public fear [35]. These apprehensions hinder the progress of AI development and can be dispelled through a comprehensive understanding of the functions outlined in this analysis.

Finally, it is necessary to revisit the Turing Test. This study does not seek to undermine it entirely but instead advocates for viewing it as a mere demonstration of AGI's capability rather than a strict intelligence criterion. AGI enthusiasts have become overly fixated on perfectly replicating Turing's original experiment. However, this narrow focus misses the broader perspective, as the thought behind the test can be extracted from the Imitation Game itself. The goal of AGI, while ambitious and perhaps even fanciful, can be generally abstracted as the ability to perceive, assimilate, and contextually understand the world in a manner akin to humans.

## 6 CONCLUSION

Time will serve as the most significant indicator for the future trajectory of AGI development. Assuming the prevailing architecture remains, the critical question becomes when these products will begin to stagnate. Once the most cutting-edge implementations reach their peak, the limitations of raw computational power will become evident, and several crucial considerations will shape the future of these applications.

The paramount question arising amidst declining performance is whether the pursuit remains a worthwhile endeavor. While this paper asserts that it is not, individual researchers must consider whether to shift their focus away from HLI. Even if they opt for a change in direction, additional considerations come into play, such as whether current attempts can be adapted to narrower contexts. If this is feasible, AGI initiatives may offer valuable insights for optimizing the learning abilities of ANI applications. Unfortunately, in the current research paradigm, the promotion of ongoing developments makes this analysis of its culmination seem impractical.

It is crucial to emphasize that this study does not seek to undermine the strength and utility of current architectures, as machine learning has revolutionized countless industries. Additionally, the challenges in AGI do not necessarily stem from the potential intelligence of the systems themselves. Instead, it resides in the capacity to efficiently assimilate and learn from vast amounts of data. As numerous studies posit, AGI is theoretically attainable, since the computational power needed to train an intelligent model is not unreasonable. However, computation alone does not encompass the contextual and experiential data necessary to train such a model. AI will excel when the focus is shifted away from the pursuit of HLI while current paradigms cannot support it. Only when it operates within an environment conducive to its strengths can AI's true potential be accurately assessed.